\pdfoutput=1

\documentclass[11pt]{article}

\usepackage{acl}

\usepackage{times}
\usepackage{latexsym}
\usepackage{graphicx}
\usepackage{colortbl}
\usepackage{multirow}
\usepackage{booktabs}
\usepackage{algorithm}
\usepackage{algpseudocode}
\usepackage{float}
\usepackage{hyperref}
\usepackage{bbm}
\usepackage{graphicx} 
\usepackage{subcaption}
\usepackage{array}
\usepackage[nolist]{acronym} 

\usepackage[T1]{fontenc}

\usepackage[utf8]{inputenc}

\usepackage{microtype}

\usepackage{inconsolata}

\title{Can OpenSource beat ChatGPT? -- A Comparative Study of Large Language Models for Text-to-Code Generation}

\author{
  \textbf{Luis Mayer\textsuperscript{1}},
  \textbf{Christian Heumann\textsuperscript{1}},
  \textbf{Matthias Aßenmacher\textsuperscript{1,2}}
\\
\\
  \textsuperscript{1}Department of Statistics, LMU Munich, Germany,\\
  \textsuperscript{2}Munich Center for Machine Learning (MCML)\\
\\
  \small{
    \textbf{Correspondence:} \href{mailto:luis.mayer@campus.lmu.de}{luis.mayer@campus.lmu.de}, \href{mailto:matthias@stat.uni-muenchen.de}{matthias@stat.uni-muenchen.de}
  }
}

\begin{document}

\maketitle

\begin{abstract}
In recent years, large language models (LLMs) have emerged as powerful tools with potential applications in various fields, including software engineering. Within the scope of this research, we evaluate five different state-of-the-art LLMs - Bard, BingChat, ChatGPT, Llama2, and Code Llama - concerning their capabilities for text-to-code generation. In an empirical study, we feed prompts with textual descriptions of coding problems sourced from the programming website LeetCode to the models with the task of creating solutions in Python. Subsequently, the quality of the generated outputs is assessed using the testing functionalities of LeetCode. The results indicate large differences in performance between the investigated models. ChatGPT can handle these typical programming challenges by far the most effectively, surpassing even code-specialized models like Code Llama. To gain further insights, we measure the runtime as well as the memory usage of the generated outputs and compared them to the other code submissions on Leetcode. A detailed error analysis, encompassing a comparison of the differences concerning correct indentation and form of the generated code as well as an assignment of the incorrectly solved tasks to certain error categories allows us to obtain a more nuanced picture of the results and potential for improvement. The results also show a clear pattern of increasingly incorrect produced code when the models are facing a lot of context in the form of longer prompts.
\end{abstract}

\section{Introduction}
\label{sec:intro}

\begin{figure}[ht]
    \centering
    \includegraphics[width=0.95\linewidth]{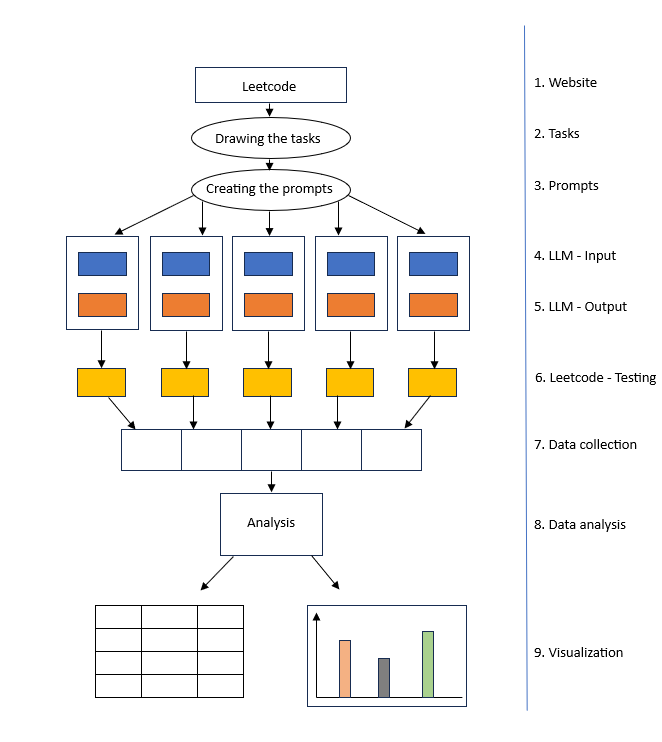}
    \caption{Schematic overview of the evaluation pipeline for the LLMs' performance on text-to-code generation.}
    \label{fig:study_process}
\end{figure}

Natural Language Processing (NLP) is an interdisciplinary field at the intersection of artificial intelligence, computer science, and linguistics. Its primary goal is to enable computers to understand, interpret, and generate human language in a valuable manner. In the early days of NLP, the focus was primarily on rule-based systems and basic statistical models \citep{zhou2020}. These systems, while innovative for their time, often struggled with constant adjustments and maintenance, linguistic variability, and scalability \citep{manning1999, jurafsky2023}. With the advent of Machine Learning, particularly Deep Learning, a paradigm shift occurred. Deep Learning models, especially since the end of the last decade, so-called large-language models (LLMs), based on enormous amounts of data and immense computational power, have demonstrated an unprecedented ability to understand and generate language \citep{zhang2021}. The emergence of LLMs is closely linked to the development of fundamental architectures such as the Transformer, introduced by \citet{vaswani17}, which fostered a new era of language models including well-known LLM-based Chatbots such as ChatGPT \cite{chatgpt}.
LLMs have proven themselves very useful for a variety of tasks ranging from answering questions, over various classification tasks to summarizing texts or writing essays. In addition to handling human language, some LLMs are also able to deal with code \cite{feng-etal-2020-codebert}. Based on textual descriptions of a programming problem as an input, these models can generate code in different programming languages as an output. The application of translating natural language into syntactically and semantically correct code has far-reaching implications, particularly in software development, where it can bridge the gap between domain experts and developers. \\
\paragraph{Contribution} With this paper, we contribute to a better understanding of recently developed LLMs in terms of their capabilities for code generation by examining them in an empirical study as depicted in Figure \ref{fig:study_process}. Within this study, we investigate the performance of five different LLMs -- ChatGPT, BingChat, Bard, Llama2, and Code Llama -- in terms of text-to-code generation for the programming language Python and compare their performance with respect to (i) correctness, (ii) runtime, and (iii) memory usage. We draw further conclusions by making use of the different difficulty levels provided in conjunction with the task formulations.

\section{Related Work}
\label{sec:related}

Studying the abilities of LLMs in coding-related tasks is a dynamically developing field of research at the moment since ongoing developments of LLMs enable their usage for a wide range of coding and programming tasks. Thus, it is crucial to distinguish between the different kinds of tasks, as their applications often differ considerably from each other. The paper of \citet{zhang2023} provides an overview of several types of code-processing tasks. In the realm of code correction, \citet{sobania2023} investigate ChatGPT's bug-fixing performance by confronting the LLM with 40 faulty Python code snippets. In a similar study \citet{zhang2023critical} examine the proficiency of ChatGPT in repairing flawed Java programs and compare its results to task-specific models like CodeT5 and PLBART. Concerning code summarization, \citet{sun2023} present an insightful study on the evaluation of ChatGPT's abilities to create comments for Python code snippets.
Likewise to the works mentioned above, also in the area of text-to-code generation, most studies conducted so far have focused on testing ChatGPT. \citet{geng2023} evaluate the performance of ChatGPT within the frame of an introductory-level functional language programming course, while \citet{piccolo2023} explore its capabilities in solving programming tasks from an introductory bioinformatics course.
The code generation abilities of ChatGPT were evaluated in another study by \citet{buscemi2023}, where the authors benchmark the model on various tasks in ten different programming languages. In the study by \citet{muennighoff2023}, the focus is on open-source LLMs that are examined in the three task areas of code correction, code explanation, and code synthesis in six different programming languages. The work of \citet{austin2021} also deals with code synthesis in Python using a wider range of LLMs. In their study, the models are employed both with and without fine-tuning and a comparative analysis of the results was conducted. A study that is somewhat similar in structure to ours, but with a greater emphasis on only whether the code is correct or not, is presented by \citet{destefanis2023}. In this work, ChatGPT and Bard are prompted to generate Java code based on a provided code description.

\section{Materials and Methods}
\label{sec:material}

\subsection{Data Set}
\label{sec:data}

The tasks for this study are sourced from LeetCode\footnote{\url{https://leetcode.com/}}, a training website providing a diverse range of programming problems, including (but not limited to) algorithms, dynamic programming, or graphs. LeetCode categorizes the tasks into different topics (e.g. array, math, or sorting) and three difficult levels (easy, medium, and hard), the latter of which we exploit to measure the LLMs' performance more granularly. Users can pick tasks and try to solve them in different programming languages. LeetCode also provides test cases to check possible solutions and several evaluation metrics for accepted solutions. The focus of this study is to test the performance of LLMs in math and statistics-related tasks. The three task topics in the LeetCode repository best suited for these constraints and most frequently represented are math, matrix, and counting. For the first two subject areas, 30 tasks are selected consisting of 10 tasks from each of the three different difficulty levels. Only 29 tasks were selected for counting, as there were only 9 tasks with a difficulty level of "hard" at the time of data collection. Across all three topics, this results in 89 tasks that are used in this study. After filtering for the appropriate types and difficulty levels, the tasks are randomly drawn using the \textit{pick one} button in LeetCode. 

\subsection{Models}
\label{sec:models}

\paragraph{ChatGPT} is an instruction-tuned LLM based on models from the GPT series \citep{radford2018improving,radford2019language,brown2020language,openai2023gpt} developed by OpenAI and initially published in November 2022. The initial and freely accessible variant of ChatGPT was based on the GPT-3.5 turbo (since March 2023), while starting from February 2023 ChatGPT paid access to a newer version based on GPT-4 has been available. For the study conducted in this work, the ChatGPT model based on the GPT-3.5 turbo version was employed, making our performance estimates for ChatGPT somewhat conservative. Further, this ChatGPT was trained on a corpus with a cut-off date in September 2021 and thus does not have access to information newer than October 2021.

\paragraph{BingChat} is a conversational LLM-based feature for Microsoft's search engine Bing that can also be used for programming tasks. So instead of typing a search query into Bing, the user can interact with BingChat \citep{xuanquy2023}. It was launched in February 2023 and is powered by the GPT-4 model from OpenAI. Due to an internet connection, BingChat has access to all the latest information and is not limited in a way like ChatGPT.

\paragraph{Bard\protect\footnote{In February 2024, Bard was renamed to Gemini.}} is the instruction-tuned variant of an LLM developed by Google. It was first released in March 2023 on a limited basis, followed by a full release in May 2023. The chatbot was at first powered by LaMDA \citep[Language Model for Dialogue Application;][]{thoppilan2022lamda}, but since May 2023 Bard has been based on Google's PaLM 2 \citep[Pathways Language Model 2;][]{anil2023}. PaLM 2 uses compute-optimal scaling to adjust the model size to the number of tokens in the pre-training corpus. This updated approach makes PaLM 2 more compact than its predecessor PaLM, while still providing higher efficiency, faster inference, and a reduced parameter count \citep{anil2023}. Similar to BingChat, Bard can access newly appearing information via the internet and is thus able to also provide links to websites and other online resources.

\paragraph{Llama2} is an open-source LLM released by Meta in July 2023. It has been trained with 40\% more data than its predecessor Llama and is capable of handling twice its context length \citep[4096 vs. 2048 tokens;][]{touvron2023}. Llama2 is offered in three versions with either 7 billion (7B), 13 billion (13B), or 70 billion (70B) parameters. For this study, we choose the largest available version (70B). The chatbot \textit{llama-2-70b-chat}, provided by the website Replicate, was employed for this purpose.

\paragraph{Code Llama} is a family of LLMs developed by Meta and published in August 2023. It's a variant of Llama2 specifically aligned for coding-related tasks, which was fine-tuned on large data sets of programming code. In fact, Code Llama is trained on the same code-specific datasets as Llama2 but using more samples from the same data set for a longer training time \citep{CodeLlama2023}.
In addition to the standard Code Llama model, there are two other versions: a Python-specialized version called Code Llama-Python and Code Llama-Instruct, which is fine-tuned for understanding prompts in natural language. All three types of Code Llama exist in a 7B, a 13B, and a 34B version \citep{rozière2023}. Since we provide the LLMs with instructions in natural language, the variant Code Llama-Instruct with the largest parameter count (34B) was selected. The chatbot \textit{Code Llama Instruct (34B)} from the website together.ai, which offers several LLMs, is used for this purpose.

\section{Experimental Setup}
\label{sec:exp}

As depicted in Figure \ref{fig:study_process}, the prompts are created based on the pre-selected tasks (cf. Sec. \ref{sec:material}). To understand the process, it is important to closely inspect their structure (cf. Appendix \ref{a:task}, Fig. \ref{fig:example_prompt}). Initially, each task consists of three parts. To create a coherent prompt, a fourth part is added and all paragraphs are separated by inserting a blank line. The first part describes the problem to be solved by a function to be created with code. In the subsequent part, one to three examples are presented to show exemplary inputs and outputs of the function. Partially these are supplemented by additional explanations. All this is complemented by a final part listing constraints and conditions that the function is required to fulfill. The fourth section we added consists of the request for the task to be solved in the programming language Python. We consistently use the expression "Write the code for this task in Python" followed by the name of the Python function with the arguments to be included since the automated tests can only be performed if the generated code uses the function names and arguments specified by LeetCode. Since this information is accessible on LeetCode for each task, we consistently use the prefix "\textit{Start with: <function\_name and arguments>}". Some of the tasks in LeetCode have integrated illustrations in the example part, which allows the user a better understanding of the problem. Since the five examined models are not capable of processing visual information, those are not included in the prompt.

After conducting all the mentioned alterations, we also kept track of the number of tokens that constituted the prompts. Then, the prompts are passed to all five models as input. For each task a new chat is started, so that a conversation with the LLM always contains exactly one prompt and one generated output per task. Since the answers often consisted of code blocks interleaved with text descriptions and exemplary applications (cf. Appendix \ref{a:ex_output}, Fig. \ref{fig:example_output}), it is crucial to locate the part that contains the function to be implemented. This part is inserted into the code field in LeetCode for the corresponding task before LeetCode's test procedure is executed. Depending on the test result, the feedback from LeetCode differs: If a task is not solved correctly, LeetCode indicates the type of error (e.g. "wrong answer" or "invalid syntax") and the number of correctly passed tests. The different error categories are described in more detail in Section \ref{sec:res}. For correctly answered tasks, LeetCode displays a new window featuring runtime, memory usage, and corresponding ranks for the code.

LeetCode measures the runtime in milliseconds and memory usage in megabytes for each successful code submission. Furthermore, these measured values are retained by Leetcode to compare them to all other runtime and memory usage values from submissions by other users, provided that the task and programming language are the same. LeetCode then calculates a quantile ranking indicating the percentage of correct submissions that a given solution surpasses in terms of performance.
As an illustration, consider a correct submitted code with runtime and memory usage rankings of 52\% and 99\%, respectively. In the context of runtime, this implies that the code is only slightly faster than half of all properly submitted codes. On the other hand, in terms of memory usage, the code outperforms nearly all other submissions, with only one percent exhibiting lower memory usage. The described metrics are stored for each task and model. The findings are presented together with visualizations in the next section.\\

\section{Results}
\label{sec:res}

\paragraph{Post-Processing of the generated outputs}
As already hinted at in the last section, the generated responses of the LLMs often consist of interleaved text and code blocks. However, our experiments show that the code does not necessarily have to be in a code block. Overall, three categories of the code location are found:

 \begin{itemize}
     \item All code in a single code block. Before/after this there may be optional text blocks. This is the standard response type for ChatGPT, Bard, and Llama2.
     \item The generated code is located in a text block. This text block might also contain descriptions and comments about the code. This case occurs sometimes with BingChat and Code Llama.
     \item The code is distributed across text and code blocks. This case occurs sometimes with BingChat and Code Llama.
 \end{itemize}

If code situated within a text block (second and third category) is tested within LeetCode, it is immediately assessed as incorrect. This arises from the structure of text blocks, where each line invariably begins with a word. Hence, the code is e.g. not indented, as it should be for instance in if-statements or for-loops as illustrated in Figures \ref{fig:example_intended} and \ref{fig:example_not_intended} (Appendix \ref{a:ex_output}). Since only due to the fact of missing indentations, otherwise correct code may not be recognized as such, we decided to intervene in such cases and to perform the necessary indentations subsequently for all affected models. This was especially crucial for Code Llama, as its code was consistently generated within text blocks. For this reason, post-processing was necessary for all 89 tasks for Code Llama. In four instances, BingChat was also affected. The adjustments were always applied after the code generation and before testing on LeetCode. The approach described here was consistently employed for all analyses and results presented in the following. 

\paragraph{Correctness}

\begin{table*}[ht]
\renewcommand{\arraystretch}{1.0}
\begin{center}
 \begin{tabular}{|c|c|c|c|c|c|}
   \hline
          & \textbf{Bard} & \textbf{BingChat} & \textbf{ChatGPT} &\textbf{ Code Llama} & \textbf{Llama2}  \\
   \hline
   \hline
   correct & 18\% (16) & 39\% (35) & 58\% (52) & 9\% (8) & 7\% (6)  \\
   
   incorrect & 82\% (73) & 61\% (54) & 42\% (37) & 91\% (81) & 93\% (83)  \\
   \hline
 \end{tabular}  
\caption{Relative (Absolute) frequencies of correct and incorrect solutions for all evaluated models.} 
\label{table:correct_incorrect_solutions}
\end{center}
\end{table*}

\begin{figure}[ht]
    \centering
    \includegraphics[width=0.95\linewidth]{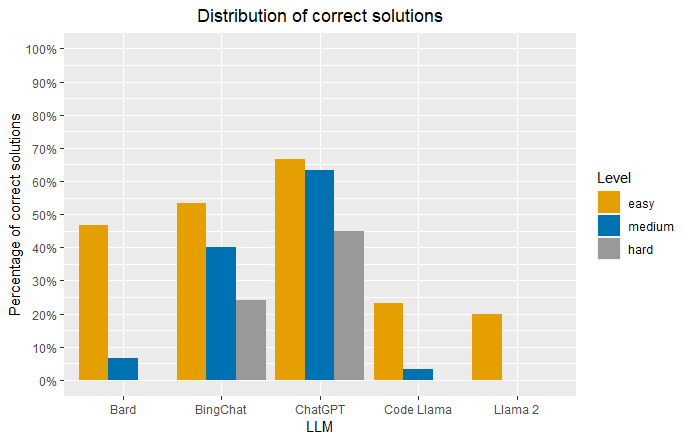}
    \caption{Percentage of correctly solved tasks per LMM.}
    \label{fig:distribution_correct}
\end{figure}

Table \ref{table:correct_incorrect_solutions} and Figure \ref{fig:distribution_correct} clearly show that the LLMs' performance differs notably. The best-performing model is ChatGPT, which is the only model to solve more than 50\% of the tasks correctly. The runner-up is the second GPT-based model, BingChat, ahead of Bard. The two Llama models solve by far the fewest tasks correctly, with a share of not even 10\%. Code Llama still performs slightly better than Llama2. Although Code Llama is based on Llama2, their output only matches in two (correctly solved) cases.

Across all evaluated LLMs we further observe (cf. Fig. \ref{fig:distribution_correct}) that the share of correctly solved tasks decreases with a higher difficulty level. While all five models are able to complete at least some of the tasks on the easy level, only four of the models achieve the right solutions on the medium level. Further, only the two GPT-based models can successfully solve any task on the highest level. While ChatGPT and BingChat consistently solve more than 20\% of tasks correctly, only Bard and Code Llama are above this mark on the easy difficulty level. At the medium difficulty level, these two LLMs drop below 10\%, corresponding to two and one correct solution, respectively.

\paragraph{Performance metrics}

In the next step, we closely inspect the correctly solved tasks. Our focus specifically lies on the runtime and memory usage metrics as explained in Section \ref{sec:exp}, which LeetCode only returns for each correct submission. As already mentioned, Bard and Code Llama only have one and two right solutions at the medium level which is why it is hardly possible to draw any conclusions and so we omit the combination of these models and difficulty levels in the following figures. It is further important to note that the following two bar plots consistently depict average values, which are composed of the rank values of the individual correctly solved tasks.

\begin{figure}[ht]
    \centering
    \includegraphics[width=0.95\linewidth]{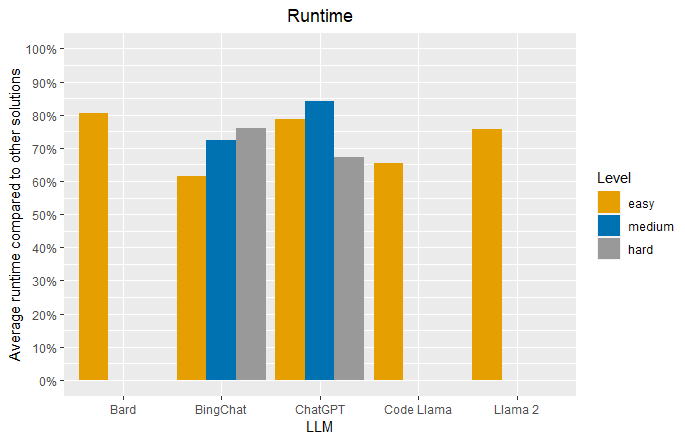}
    \caption{Average runtime ranking (quantiles) of the generated code per LLM and difficulty level.}
    \label{fig:distribution_runtime}
\end{figure}

In Figure \ref{fig:distribution_runtime} the average of the runtime ranks per LLM and difficulty level is depicted. For a better understanding of the plot, consider the bar on the left side: Bard solved 14 tasks correctly for the difficulty level "easy", so Leetcode also calculated 14 ranking values for the runtime, the average of which is displayed here. A value of 70\% means that the average runtime of the generated code beats 70\% of all other code submissions for this task. Conversely, only 30\% of the submitted solutions are quicker. It is noticeable that the (correct) LLM-generated solutions exhibit a comparatively low average runtime, as they are ranked above 50\% on average. While concerning the number of correct solutions there are clear differences between the models and the difficulty levels, this is not the case for the runtime: A decrease in runtime performance with higher difficulty levels is not evident. It can be argued that ChatGPT outperforms the other models concerning the number of correctly solved tasks, but not (consistently) in terms of runtime. The ranking values for all models lie between 61\% (BingChat, "easy") and 84\% (ChatGPT, "medium"). These values suggest that the runtime of the code produced by the models is often lower than that of other code submissions. As faster code is favorable in the context of software development, this is an aspect in which LLMs might be able to contribute effectively.

\begin{figure}[ht]
    \centering
    \includegraphics[width=0.95\linewidth]{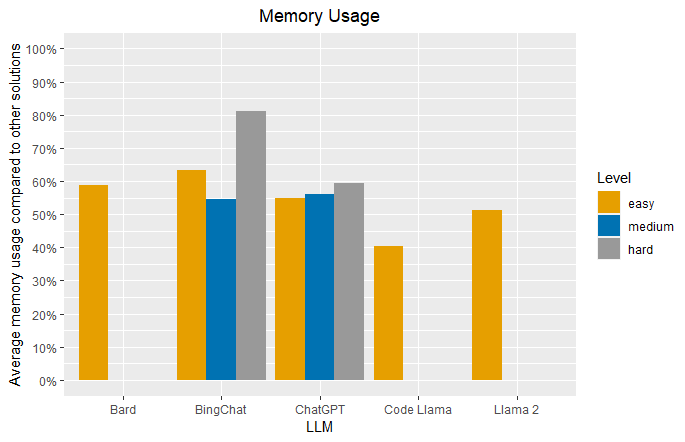}
    \caption{Average memory usage ranking (quantiles) of the generated code per LLM and difficulty level.}
    \label{fig:distribution_memory}
\end{figure}

Concerning memory usage, the average values per model and difficulty level are calculated and displayed in Figure \ref{fig:distribution_memory} according to the same principles as for the runtime. A higher value signifies that the code generated by the models utilizes less memory than solutions provided by other users. Reduced memory usage is advantageous as it allows for easier scalability to handle larger data volumes without additional memory requirements. Besides, code with lower memory usage is often more efficient, especially when required to operate on systems with limited RAM. The main takeaway from taking this angle is that the differences between the models are (again) not as pronounced as observed for correctness. While ChatGPT only achieves similarly high values as the other models, BingChat performs notably better across difficulty levels. Except for Code Llama, all values are again above 50\%, although the values are mostly a bit lower than those for the runtime. Exceptional performance can be reported for BingChat at the "hard" level with 81\%. 

\begin{figure}[ht]
    \centering
    \includegraphics[width=0.95\linewidth]{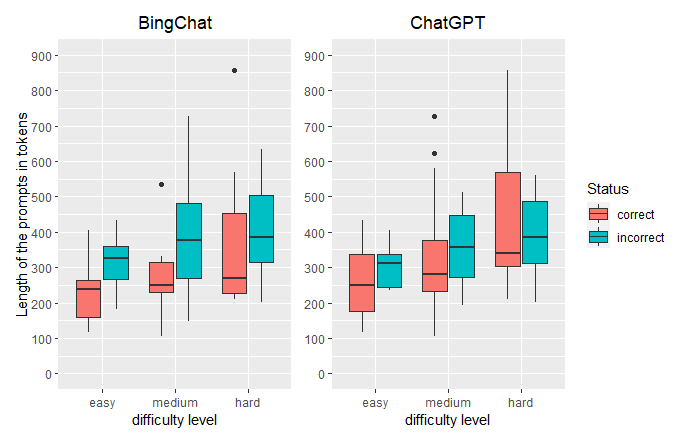}
    \caption{Distribution of prompt lengths of correct and incorrect solutions for BingChat and ChatGPT}
    \label{fig:prompt_lengths_GPT}
\end{figure}

\paragraph{Further insights and error analysis}
Figure \ref{fig:prompt_lengths_GPT} presents the distribution of prompt lengths of correct and incorrect solutions, measured by the number of tokens the prompts consist of. For our main analysis, we focus solely on BingChat and ChatGPT since they are the best-performing LLMs regarding the number of correct solutions and at the same time the only ones with correct solutions at all difficulty levels. The boxplots for the remaining three LLMs can be found in Figure \ref{fig:prompt_lengths_Bard_Llama_Python} in Appendix \ref{a:lengths}. We observe that the lengths of prompts are smaller for correct solutions than for incorrect solutions at all three difficulty levels. This conversely seems to imply that shorter prompts are positively associated with the likelihood of the models producing a correct solution.

\begin{table*}[ht]
\renewcommand{\arraystretch}{0.85}
\begin{center}
 \begin{tabular}{|c|c|c|c|}
 \hline
   & \textbf{Error type} & \textbf{Count}  & \textbf{Share in \%  of all errors} \\
   \hline
   \hline
   1 & wrong answer & 178  & 54.3\%\\ 
   \hline
   2 & syntax error & 48  & 14.6\%\\
   \hline
   3 & type error & 18  & 5.5\%\\
   \hline
   4 & name error & 17  & 5.2\%\\
   \hline
   5 & time limit exceeded & 15  & 4.6\%\\
   \hline
   6 & indentation error & 10  & 3.0\%\\
   \hline
   7 & index error & 10  & 3.0\%\\
   \hline
   8 & attribute error & 9  & 2.7\%\\
   \hline
   9 &  no code generated & 7  & 2.1\%\\
   \hline
   10 & zero division error & 4  & 1.2\%\\
   \hline
 \end{tabular}  
\caption{Ten most frequent error categories among all LLMs} 
\label{table:top_ten_error_categories}
\end{center}
\end{table*}

\begin{figure}[ht]
    \centering
    \includegraphics[width=0.95\linewidth]{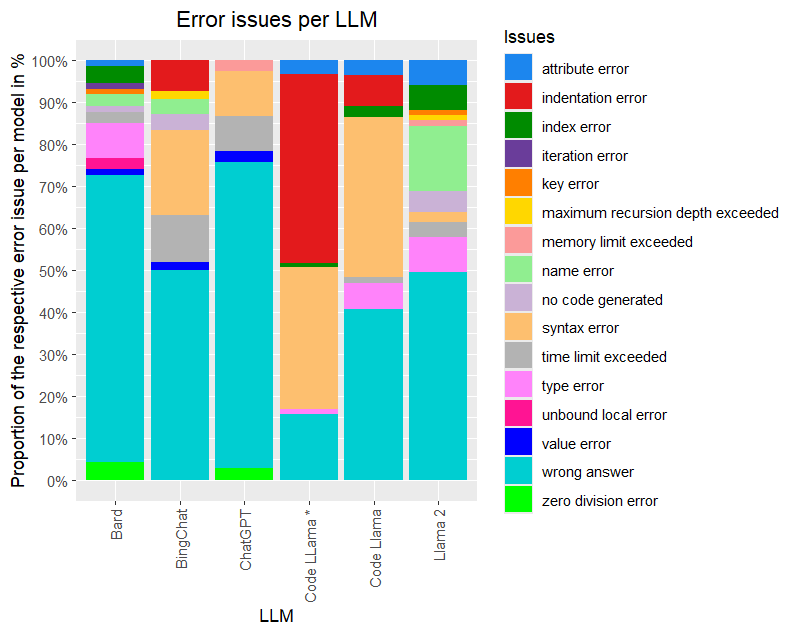}
    \caption{All error categories per LLM.}
    \label{fig:all_error_issues_percent}
\end{figure}

Concluding the analysis, Table \ref{table:top_ten_error_categories} summarizes the ten most frequent errors across all five models. A complete enumeration of all errors can be found in Table \ref{table:all_errror_categories} in Appendix \ref{a:error}. We learn from this table that the error category \textit{"wrong answer"} is by far the most frequent one with a share of $>50$\%. If this error is encountered, this indicates that the code submission has passed only a specific number of tests in LeetCode, which is less than the total number of tests required for a correct solution. Therefore, the generated outputs affected by \textit{"wrong answer"} do not cover the required functionality expected from the code. The second most common reason why solutions from LeetCode were rejected is \textit{"syntax error"} with approximately 15\%. All other error categories depicted in Table \ref{table:top_ten_error_categories} are clearly below 10\%. The category \textit{"type error"} (5.5\%) occurs when a function or operation is applied to an object of an inappropriate type, such as attempting to concatenate an object of type \texttt{integer} with an object of type \texttt{string}. If undefined variables are used in the code, the error message falls into the category \textit{"name error"} (5.2\%). The error message \textit{"time limit exceeded"} (4.6\%) means that LeetCode cannot perform all tests, potentially due to an infinite loop in the generated code or simply because the code execution time exceeds LeetCode's time limit. An \textit{"index error"} (3\%) appears when trying to access an index that is outside the bounds of a sequence type, such as arrays or lists. \textit{"Attribute errors"} (2.7\%) happen when accessing attributes of an object that it is not ascribed. The error category \textit{"no code generated}", which means that the model did not return any code as output to the prompt, was only the ninth most frequent error with seven cases in total (2.1\%). These seven cases are divided among Llama2 (four times), BingChat (two times), and Bard (only once). As the name already implies, a \textit{"zero division error"} (1.2\%) happens when attempting to execute a division operation where the divisor is zero. Figure \ref{fig:all_error_issues_percent} displays all error categories that occurred per model. The height indicates the share of the error category (in \%) of all incorrectly solved tasks per model. For comparison purposes, we also report the results of Code Llama when the missing indentations are not corrected. These cases are denoted as Code Llama* and shown next to the other models.

It is striking that the category \textit{"wrong answer"} is the only one in the top three for all LLMs. Except for Code Llama*, it is the most common reason for errors as is to be expected from the results presented in Table \ref{table:top_ten_error_categories}. For Bard and both GPT-based models, there is the highest margin between this category and the runner-ups. We also observe that BingChat and ChatGPT do not differ concerning the three most frequent error categories and the order in which they appear.

If we now compare the two Code Llama approaches to each other, i.e. the one with added indentations (that was also used in the rest of our analysis) to the unaltered approach (in which we test the generated code without adaptation), clear differences are noticeable in the most frequent error categories. While the \textit{"indentation error"} is in first place in the unaltered approach, it is only in third place after adaptation. As this error category has decreased, the number of incorrectly solved tasks due to \textit{"wrong answer"} rises. The error category \textit{"syntax error"}, however, remains almost constant for both approaches.

\section{Conclusion}
\label{sec:concl}

We conducted an empirical study on five different LLMs, namely BingChat, ChatGPT, Bard, Llama2, and Code Llama, aiming at evaluating their potential for text-to-code generation. The mentioned LLMs were examined on 89 Python programming tasks from the coding website LeetCode. The results show clear differences between models and the number of code outputs they could generate correctly. In conclusion, ChatGPT outperformed all other models by a notable margin, followed by the second GPT-based model BingChat. Code Llama and Llama2 exhibited the lowest proficiency in solving tasks correctly, with Code Llama only surpassing Llama2 by correctly solving two additional tasks. Consequently, it cannot be affirmed that Code Llama, despite being a model specialized in coding-related tasks, significantly outperforms its foundational model Llama2 in text-to-code generation. Moreover, although all the models were able to produce code in Python, a notable drawback of Code Llama is that the generated code is not indented, requiring the user to adjust the format of the code before it can be effectively deployed. Upon examining the correct solutions for all tasks, it can be stated that the code outputs generated by the LLMs are in many cases more efficient in terms of runtime and memory usage than human-written code solutions, assuming that almost all submissions on LeetCode are from humans.
Regardless of the model, the length of the prompts seems to have an apparent influence on the likelihood of generating correct code. Since the prompts for correct solutions were shorter, it can be tentatively concluded that the prompts should be designed in a brief and targeted manner to achieve better results.

Moreover, it is worth noting that the models' failures stem to a large extent from code errors, categorized as \textit{"wrong answer"}, rather than due to supposedly simpler syntax errors. This is especially the case for Bard, BingChat, and ChatGPT. Considering the definition of the error category \textit{"wrong answer"} as outlined in Section \ref{sec:res}, it can be concluded that the three models might even demonstrate partial correctness in solving many tasks. Even these partially correct code outputs can be helpful for the user in subsequent applications, as incorporating LLMs has already advanced them closer to a solution. Users can then continue to work on the correct solution based on the generated code outputs. The outcomes indicate that ChatGPT, in particular, emerges as a useful assistant in terms of text-to-code generation. The impression gained from our study conducted is largely consistent with the findings of \citet{geng2023}. Additionally, the observed strong performance of ChatGPT is also evident in a study by \citet{buscemi2023}, in which ChatGPT had to generate code in ten different programming languages on the same tasks. In the same study, the authors concluded that the model performed well in Python. \citet{buscemi2023} speculates on a possible reason that ChatGPT performs best in the programming languages in which it has received the most training signals. According to ChatGPT itself, Python is one of the top ten languages on which it has been trained. Nevertheless, considering the large number of wrong solutions, one should always look critically at the generated code, which still requires substantial knowledge of the programming language by the user.

While in the study the conversation between user and model consists only of one input and output, more extensive dialogues should be explored in future experiments. These dialogues could involve instructing the model to identify and fix errors in the previously generated code. Alternatively, they might encompass the execution of several related programming tasks in a single conversation, which requires the model to access previously generated code. Future studies can be extended to other programming languages or task types.

\section*{Ethical Considerations and Limitations}

It is important to interpret the results with caution, as several limitations need to be considered. First, it is crucial to point out that the study conducted is only a snapshot of the performance of LLMs within the field of code generation at a certain point in time. The field of LLMs continues to move extremely rapidly, and new models and architectures may already outperform previous ones \citep{zhang2021}. Second, the study encompasses only 89 tasks and is thus limited in its scope. This has a direct influence on the interpretation and the generalizability of the results. The task volume affects the explanatory power and statistical robustness of the findings, with outcomes derived from a more comprehensive task set naturally having greater evidential weight. Third, we do not claim that our study is exhaustive, so as presented in Section \ref{sec:related} there are several other code-related tasks that LLMs can implement besides text-to-code generation. In addition, the scope of the work was also restricted to Python, which limits the applicability of the results to other programming languages. In terms of content, difficulty, and length, the selected tasks only cover a small subset of the huge spectrum of code-related tasks that can be passed to LLMs. Therefore, the tasks used do not embody the full bandwidth of challenges associated with text-to-code generation tasks, constraining the expressiveness of the study results to a narrow and focused set of tasks. 

Concerning the reproducibility of the results, it is essential to acknowledge that the comparison values used in the runtime and memory usage analyses will change over time since users always submit new solutions to LeetCode. Therefore, these findings should rather be considered as a snapshot. When submitting the LLM-generated results to LeetCode, it was not possible to flag them as such or to prevent them from being incorporated into the rankings. However, given the small number of solutions we uploaded, we do not think this will notably influence the LeetCode statistics. Furthermore, the reproducibility of results in the study is influenced by the fact that LLMs produce different answers even to identical prompts as inputs. This occurs because the models are partially updated and incorporate a certain degree of randomness in their responses. The generated outputs may differ from those obtained in the study conducted.

\section*{Acknowledgements}
This work has been partially funded by the Deutsche Forschungsgemeinschaft (DFG, German Research Foundation) as part of BERD@NFDI - grant number 460037581.

\bibliography{custom}

\clearpage

\appendix

\section{Leetcode tasks}
\label{a:task}

\begingroup
\let\clearpage\relax 
\onecolumn
\endgroup

\begin{figure}[ht]
    \centering
    \includegraphics[width=0.95\linewidth]{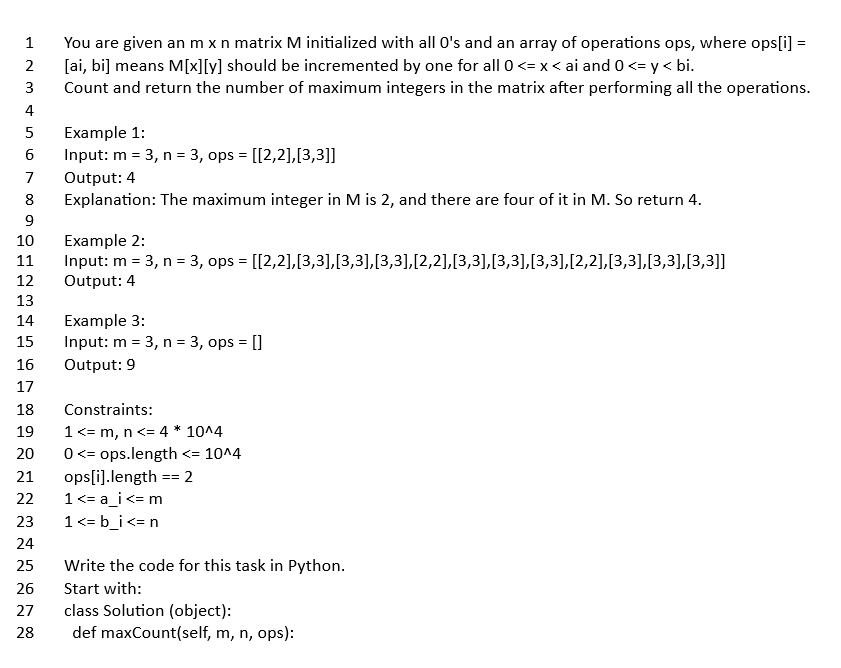}
    \caption{Examplary prompt based on LeetCode task number 598: Range Addition 2}
    \label{fig:example_prompt}
\end{figure}

\clearpage

\section{Exemplary Output}
\label{a:ex_output}

\begin{figure}[ht]
    \centering
    \includegraphics[width=0.8\linewidth]{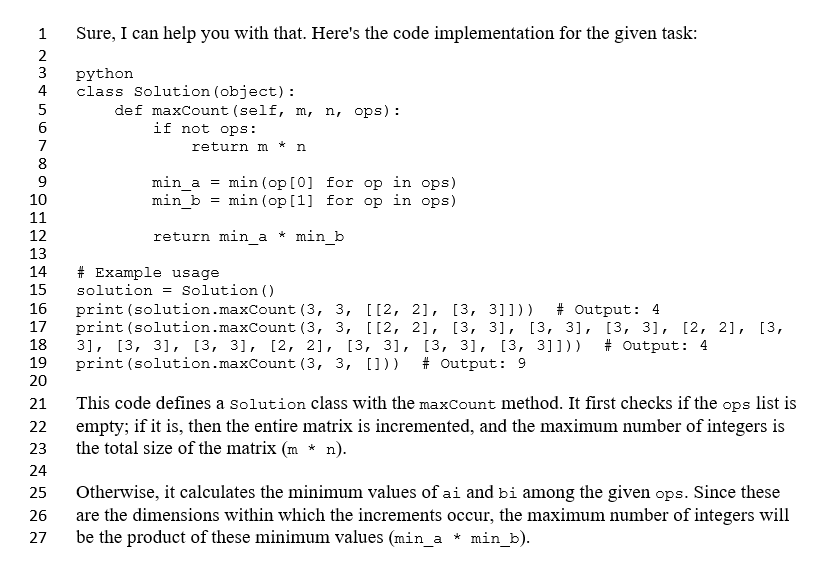}
    \caption{Output generated by ChatGPT to the prompt based on LeetCode task number 598: Range Addition 2}
    \label{fig:example_output}
\end{figure}

\begin{figure}[ht]
    \centering
    \includegraphics[width=0.8\linewidth]{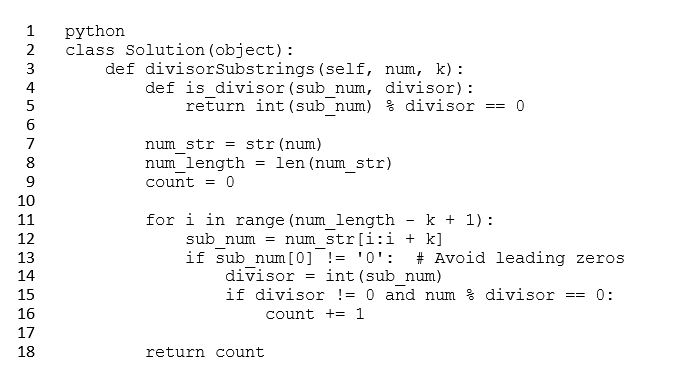}
    \caption{Indented code in a code field generated by ChatGPT (based on LeetCode task number 2269: Find the K-Beauty of a Number)}
    \label{fig:example_intended}
\end{figure}

\begin{figure}[ht]
    \centering
    \includegraphics[width=0.9\linewidth]{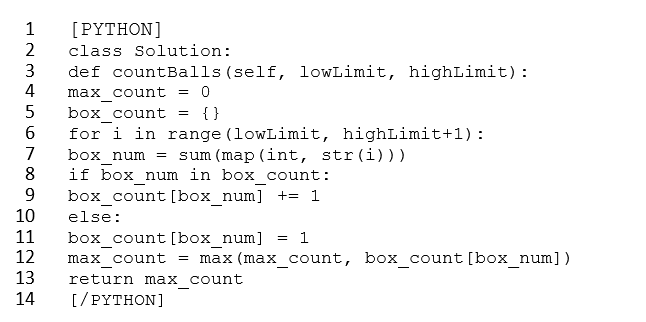}
    \caption{Non-indented code in a text field generated by Code Llama (based on LeetCode task number 2269: Find the K-Beauty of a Number)}
    \label{fig:example_not_intended}
\end{figure}

\clearpage

\section{Prompt Lengths}
\label{a:lengths}

\begin{figure}[ht]
    \centering
    \includegraphics[width=0.95\linewidth]{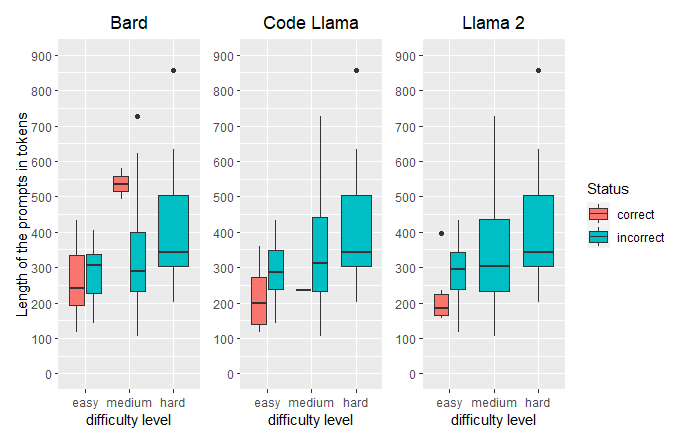}
    \caption{Distribution of prompt lengths of correct and incorrect solutions in Python for Bard, Code Llama, and Llama2}
    \label{fig:prompt_lengths_Bard_Llama_Python}
\end{figure}

\clearpage

\section{Error Analysis}
\label{a:error}

\begin{table}[ht]
\renewcommand{\arraystretch}{1.5}
\begin{center}
 \begin{tabular}{|c|c|c|c|}
 \hline
   & \textbf{Error type} & \textbf{Count}  & \textbf{Share in \%  of all errors} \\
   \hline
   \hline
   1 & wrong answer & 178  & 54.3\%\\ 
   \hline
   2 & syntax error & 48  & 14.6\%\\
   \hline
   3 & type error & 18  & 5.5\%\\
   \hline
   4 & name error & 17  & 5.2\%\\
   \hline
   5 & time limit exceeded & 15  & 4.6\%\\
   \hline
   6 & indentation error & 10  & 3.0\%\\
   \hline
   7 & index error & 10  & 3.0\%\\
   \hline
   8 & attribute error & 9  & 2.7\%\\
   \hline
   9 &  no code generated & 7  & 2.1\%\\
   \hline
   10 & zero division error & 4  & 1.2\%\\
   \hline
   11 & value error & 3 & 0.9\%\\
   \hline
   12 & key error & 2  & 0.6\%\\
   \hline
   13 & maximum recursion depth exceeded & 2 & 0.6\%\\
   \hline
   14 & memory limit exceeded & 2 & 0.6\%\\
   \hline
   15 & unbound local error & 2 & 0.6\%\\
   \hline
   16 & iteration error & 1 & 0.3\%\\
   \hline
 \end{tabular}  
\caption{All occurred error categories and their frequencies} 
\label{table:all_errror_categories}
\end{center}
\end{table}

\clearpage

\section{Correctly Solved Tasks}
\label{a:correct}

\begin{table}[ht]
\setlength{\tabcolsep}{8pt}
\renewcommand{\arraystretch}{1.5}
\begin{center}
 \begin{tabular}{|c|c|c|c|c|c|c|}
 \hline
 \textbf{ LLM} &\textbf{ Level} & \textbf{Math}  & \textbf{Matrix} & \textbf{Counting} & \textbf{Total} & \textbf{Total in percent}  \\
   \hline
   \hline
   Bard & easy & 3  & 5 & 6 & 14 & 46.67\%\\
   \hline
   Bard & medium & 1 & 1 & 0 & 2 & 6.67\%\\
   \hline
   Bard & hard & 0 & 0 & 0 & 0 & 0.00\%\\
   \hline
   \hline
   BingChat & easy & 5  & 5 & 6 & 16 & 53.33\%\\
   \hline
   BingChat & medium & 4 & 5 & 3 & 12 & 40.00\%\\
   \hline
   BingChat & hard & 1 & 5 & 1 & 7 & 24.14\%\\
   \hline
   \hline
   ChatGPT & easy & 7  & 6 & 7 & 20 & 66.67\%\\
   \hline
   ChatGPT & medium & 6 & 7 & 6 & 19 & 63.33\%\\
   \hline
   ChatGPT & hard & 6 & 5 & 2 & 13 & 44.83\%\\
   \hline
   \hline
   Code Llama & easy & 2  & 2 & 3 & 7 & 23.33\%\\
   \hline
   Code Llama & medium & 1 & 0 & 0 & 1 & 3.33\%\\
   \hline
   Code Llama & hard & 0 & 0 & 0 & 0 & 0.00\%\\
   \hline
   \hline
   Llama2 & easy & 1  & 2 & 3 & 6 & 20.00\%\\
   \hline
   Llama2 & medium & 0 & 0 & 0 & 0 & 0.00\%\\
   \hline
   Llama2 & hard & 0 & 0 & 0 & 0 & 0.00\%\\
   \hline
   \end{tabular}  
\caption{Distribution of correctly solved tasks in the study with Python (the levels easy and difficult consist of 30 tasks each and difficult of 29 tasks)} 
\label{table:6}
\end{center}
\end{table}

\begin{table}[ht]
\setlength{\tabcolsep}{8pt}
\renewcommand{\arraystretch}{1.5}
\begin{center}
 \begin{tabular}{|c|c|c|c|}
 \hline
 \textbf{ LLM }& \textbf{Level} &  \textbf{Average runtime ranking} & \textbf{Average memory usage}  \\
   \hline
   \hline
   Bard & easy &  80.39\% & 58.87\%\\
   \hline
   Bard & medium & 80.46\% & 66.64\%\\
   \hline
   Bard & hard & 0.00\% & 0.00\%\\
   \hline
   \hline
   BingChat & easy & 61.40\%  & 63.38\%\\
   \hline
   BingChat & medium & 72.45\% & 54.70\%\\
   \hline
   BingChat & hard & 75.85\% & 80.98\%\\
   \hline
   \hline
   ChatGPT & easy & 78.64\%  & 54.86\%\\
   \hline
   ChatGPT & medium & 84.25\% & 56.19\%\\
   \hline
   ChatGPT & hard & 67.35\% & 59.31\%\\
   \hline
   \hline
   Code Llama & easy & 65.28\%  & 40.30\%\\
   \hline
   Code Llama & medium & 96.32\% & 98.42\%\\
   \hline
   Code Llama & hard & 0.00\% & 0.00\%\\
   \hline
   \hline
   Llama2 & easy & 75.59\%  &  51.34\%\\
   \hline
   Llama2 & medium & 0.00\% & 0.00\%\\
   \hline
   Llama2 & hard & 0.00\% & 0.00\%\\
   \hline
   \end{tabular}  
\caption{Average runtime and memory usage ranking per LLM and Level (corresponds to the plots \ref{fig:distribution_runtime} and \ref{fig:distribution_memory})}
\label{table:7}
\end{center}
\end{table}

\clearpage

\section{Absolute values for the runtime and memory usage}
\label{a:abs_run}

\begin{table}[ht]
\setlength{\tabcolsep}{8pt}
\begin{center}
 \begin{tabular}{|*{11}{c|}}
 \hline
\textbf{easy}    & \multicolumn{2}{c|}{\textbf{Bard}}
            & \multicolumn{2}{c|}{\textbf{BingChat}}
                    & \multicolumn{2}{c|}{\textbf{ChatGPT}}
                        & \multicolumn{2}{c|}{\textbf{Code Llama}}    & \multicolumn{2}{c|}{\textbf{Llama 2}}   \\
 \hline                           
 \textbf{ID} & \textbf{ms} & \textbf{mb} & \textbf{ms} & \textbf{mb} & \textbf{ms} & \textbf{mb} & \textbf{ms} & \textbf{mb} & \textbf{ms} & \textbf{mb}\\
   \hline
   \hline
   598 & - & - & 39 & 15.48 & 53 & 15.44 & - & - & - & -\\
   \hline
   1863 & - & - & 87 & 13.27 & - & - & - & - & - & -\\
   \hline
   2409 & - & - & - & - & 6 & 13.31 & - & - & - & -\\
   \hline
   2269 & - & - & - & - & - & - & - & - & - & -\\
   \hline
   441 & - & - & 435 & 13.06 & 24 & 13.22 & - & - & - & -\\
   \hline
   1742 & 282 & 16.17 & - & - & 313 & 15.94 & 858 & 15.94 & - & -\\
   \hline
   2525 & 14 & 13.32 & - & - & 12 & 13.29 & - & - & - & -\\
   \hline
   2119 & - & - & 19 & 13,33 & 19 & 13.24 & - & - & 13 & 13,14\\
   \hline
   231 & 20 & 13.37 & 22 & 13.16 & 9 & 13.26 & 18 & 13.17 & - & -\\
   \hline
   2591 & - & - & - & - & - & - & - & - & - & -\\
   \hline
   \hline
   1380 & 93 & 13.63 & 86 & 13.70 & 101 & 13.47 & - & - & - & -\\
   \hline
   1260 & - & - & - & - & 111 & 13.64 & - & - & - & -\\
   \hline
   733 &  48 & 13.43 & 47 & 13.49 & 52 & 13.56 & - & - & - & -\\
   \hline
   2500 & - & - & - & - & - & - & - & - & - & -\\
   \hline
   2643 & - & - & - & - & - & - & 815 & 13.78 & - & -\\
   \hline
   1030 & 103 & 15.85 & - & - & 113 & 16.66 & - & - & 101 & 15.81\\
   \hline
   2091 & - & - & - & - & - & - & - & - & - & -\\
   \hline
   1351 & - & - & 94 & 14.36 & 86 & 14.37 & 88 & 14.50 & 92 & 14.55\\
   \hline
   1886 & 30 & 13.25 & 68 & 24.58 & 23 & 13.25 & - & - & - & -\\
   \hline
   2319 & 198 & 14.28 & 197 & 14.24 & - & - & - & - & - & -\\
   \hline
   \hline
   1512 & 16 & 13.14 & 26 & 15.48 & 14 & 13.33 & - & - & 16 & 13.14\\
   \hline
   1897 & 31 & 13.57 & 45 & 13.27 & 32 & 13.74 & - & - & - & -\\
   \hline
   1370 & - & - & - & - & - & - & - & - & - & -\\
   \hline
   1356 & - & - & - & - & 52 & 13.53 & 49 & 13.53 & - & -\\
   \hline
   2423 & - & - & - & - & - & - & - & - & - & -\\
   \hline
   2351 & 12 & 13.26 & 10 & 13.30 & - & - & 4 & 13.36 & 10 & 13.36\\
   \hline
   1876 & 15 & 13.27 & 20 & 13.24 & 24 & 13.35 & - & - & - & -\\
   \hline
   2053 & - & - & 47 & 13.33 & 42 & 13.67 & - & - & - & -\\
   \hline
   1941 & 21 & 13.12 & - & - & 14 & 13.44 & - & - & 24 & 13.40\\
   \hline
   169 & 116 & 14.9 & 141 & 14.89 & 122 & 14.87 & 129 & 15.02 & - & -\\
   \hline
\end{tabular}  
\caption{Absolute values for runtime in milliseconds (ms) and memory usage in megabytes (mb) for each LLM and Level easy (column ID corresponds to the task number in LeetCode)}
\label{table:8}
\end{center}
\end{table}

\begin{table}[ht]
\setlength{\tabcolsep}{8pt}
\begin{center}
 \begin{tabular}{|*{11}{c|}}
 \hline
\textbf{medium}     & \multicolumn{2}{c|}{\textbf{Bard}}
            & \multicolumn{2}{c|}{\textbf{BingChat}}
                    & \multicolumn{2}{c|}{\textbf{ChatGPT}}
                        & \multicolumn{2}{c|}{\textbf{Code Llama}}    & \multicolumn{2}{c|}{\textbf{Llama 2}}   \\
 \hline                           
 \textbf{ID} & \textbf{ms} & \textbf{mb} & \textbf{ms} & \textbf{mb} & \textbf{ms} & \textbf{mb} & \textbf{ms} & \textbf{mb} & \textbf{ms} & \textbf{mb}\\
   \hline
   \hline
   2579 & - & - & 122 & 16.42 & - & - & - & - & - & -\\
   \hline
   1017 & - & - & - & - & 12 & 13.28 & - & - & - & -\\
   \hline
   319 & - & - & 16 & 13.22 & 16 & 13.43 & 6 & 13.07 & - & -\\
   \hline
   523 & - & - & 781 & 33.40 & 745 & 33.46 & - & - & - & -\\
   \hline
   2745 & - & - & - & - & 22 & 13.41 & - & - & - & -\\
   \hline
   497 & 154 & 17.22 & - & - & - & - & - & - & - & -\\
   \hline
   963 & - & - & - & - & - & - & - & - & - & -\\
   \hline
   150 & - & - & - & - & 29 & 15.25 & - & - & - & -\\
   \hline
   2063 & - & - & - & - & - & - & - & - & - & -\\
   \hline
   478 & - & - & 93 & 24.76 & 104 & 24.90 & - & - & - & -\\
   \hline
   \hline
   861 & - & - & - & - & - & - & - & - & - & -\\
   \hline
   427 & - & - & - & - & 86 & 15.34 & - & - & - & -\\
   \hline
   1605 & - & - & 709 & 18.52 & 672 & 17.74 & - & - & - & -\\
   \hline
   2684 & - & - & 1034 & 24.46 & 1386 & 22.74 & - & - & - & -\\
   \hline
   2711 & - & - & - & - & - & - & - & - & - & -\\
   \hline
   1895 & - & - & 3247 & 13.39 & - & - & - & - & - & -\\
   \hline
   2482 & - & - & - & - & 1201 & 56.36 & - & - & - & -\\
   \hline
   1536 & - & - & 412 & 14.01 & 398 & 13.75 & - & - & - & -\\
   \hline
   1926 & 581 & 15.39 & - & - & 591 & 16.94 & - & - & - & -\\
   \hline
   934 & - & - & 292 & 14.57 & 310 & 16.67 & - & - & - & -\\
   \hline
   \hline
   1497 & - & - & - & - & 44 & 25.38 & - & - & - & -\\
   \hline
   1519 & - & - & 2482 & 180.85 & 1600 & 183.93 & - & - & - & -\\
   \hline
   1010 & - & - & 242 & 16.55 & 205 & 16.51 & - & - & - & -\\
   \hline
   2182 & - & - & - & - & - & - & - & - & - & -\\
   \hline
   869 & - & - & 10 & 13.10 & 17 & 13.20 & - & - & - & -\\
   \hline
   945 & - & - & - & - & 648 & 24.42 & - & - & - & -\\
   \hline
   811 & - & - & - & - & - & - & - & - & - & -\\
   \hline
   2170 & - & - & - & - & - & - & - & - & - & -\\
   \hline
   1267 & - & - & - & - & 385 & 14.55 & - & - & - & -\\
   \hline
   1775 & - & - & - & - & - & - & - & - & - & -\\
   \hline
\end{tabular}  
\caption{Absolute values for runtime in milliseconds (ms) and memory usage in megabytes (mb) for each LLM and Level medium (column ID corresponds to the task number in LeetCode)}
\label{table:9}
\end{center}
\end{table}

\begin{table}[ht]
\setlength{\tabcolsep}{8pt}
\begin{center}
 \begin{tabular}{|*{11}{c|}}
 \hline
\textbf{hard}    & \multicolumn{2}{c|}{\textbf{Bard}}
            & \multicolumn{2}{c|}{\textbf{BingChat}}
                    & \multicolumn{2}{c|}{\textbf{ChatGPT}}
                        & \multicolumn{2}{c|}{\textbf{Code Llama}}    & \multicolumn{2}{c|}{\textbf{Llama 2}}   \\
 \hline                           
 \textbf{ID} & \textbf{ms} & \textbf{mb} & \textbf{ms} & \textbf{mb} & \textbf{ms} & \textbf{mb} & \textbf{ms} & \textbf{mb} & \textbf{ms} & \textbf{mb}\\
   \hline
   \hline
   2019 & - & - & - & - & 1791 & 15.30 & - & - & - & -\\
   \hline
   2584 & - & - & - & - & - & - & - & - & - & -\\
   \hline
   2338 & - & - & - & - & - & - & - & - & - & -\\
   \hline
   1835 & - & - & - & - & 681 & 25.84 & - & - & - & -\\
   \hline
   1735 & - & - & - & - & 757 & 19.28 & - & - & - & -\\
   \hline
   964 & - & - & - & - & - & - & - & - & - & -\\
   \hline
   1467 & - & - & - & - & - & - & - & - & - & -\\
   \hline
   2197 & - & - & - & - & 4131 & 27.74 & - & - & - & -\\
   \hline
   1510 & - & - & 699 & 16.70 & 696 & 16.74 & - & - & - & -\\
   \hline
   381 & - & - & - & - & 352 & 70.51 & - & - & - & -\\
   \hline
   \hline
   1074 & - & - & - & - & - & - & - & - & - & -\\
   \hline
   212 & - & - & - & - & 7963 & 14.97 & - & - & - & -\\
   \hline
   773 &  - & - & - & - & 47 & 13.24 & - & - & - & -\\
   \hline
   782 & - & - & 49 & 13.21 & - & - & - & - & - & -\\
   \hline
   1970 & - & - & - & - & - & - & - & - & - & -\\
   \hline
   2577 & - & - & - & - & - & - & - & - & - & -\\
   \hline
   980 & - & - & 33 & 13.18 & 40 & 13.46 & - & - & - & -\\
   \hline
   37 & - & - & 442 & 13.31 & 413 & 13.36 & - & - & - & -\\
   \hline
   1293 & - & - & 49 & 14.74 & - & - & - & - & - & -\\
   \hline
   827 & - & - & 2420 & 22.47 & 1935 & 22.67 & - & - & - & -\\
   \hline
   \hline
   2499& - & - & - & - & - & - & - & - & - & -\\
   \hline
   1857 & - & - & - & - & 2255 & 82.90 & - & - & - & -\\
   \hline
   2014 & - & - & - & - & - & - & - & - & - & -\\
   \hline
   2547 & - & - & - & - & - & - & - & - & - & -\\
   \hline
   992 & - & - & 342 & 15.17 & 336 & 16.36 & - & - & - & -\\
   \hline
   2025 & - & - & - & - & - & - & - & - & - & -\\
   \hline
   2416 & - & - & - & - & - & - & - & - & - & -\\
   \hline
   2514 & - & - & - & - & - & - & - & - & - & -\\
   \hline
   1819 & - & - & - & - & - & - & - & - & - & -\\
   \hline
\end{tabular}  
\caption{Absolute values for runtime in milliseconds (ms) and memory usage in megabytes (mb) for each LLM and Level hard (column ID corresponds to the task number in LeetCode)}
\label{table:10}
\end{center}
\end{table}

\end{document}